\documentclass[a4paper, 10pt,twocolumn]{article}
\usepackage{latex12}
\usepackage{times}
\usepackage{balance}

\usepackage{amssymb}
\usepackage{graphicx}
\usepackage{algorithmic}

\usepackage{algorithm}
\usepackage{algorithmic}

\usepackage{amsfonts}
\usepackage{subfigure}
\usepackage{multirow}
\usepackage{booktabs}
\usepackage{tabularx}
\usepackage{multirow}
\usepackage{balance}

\begin{document}

\title{Ensemble-Driven Support Vector Clustering: \\From Ensemble Learning to Automatic Parameter Estimation}

\author{Dong Huang$^1$, Chang-Dong Wang$^{2,3,4}$, Jian-Huang Lai$^{2,3}$, Yun Liang$^1$, Shan Bian$^1$, Yu Chen$^1$\\
\emph{$^1$College of Mathematics and Informatics, South China Agricultural University, Guangzhou, China}\\
\emph{$^2$School of Data and Computer Science, Sun Yat-sen University, Guangzhou, China}\\
\emph{$^3$Guangdong Key Laboratory of Information Security Technology, Guangzhou, China}\\
\emph{$^4$Key Laboratory of Machine Intelligence and Advanced Computing, Ministry of Education, China}\\
huangdonghere@gmail.com, changdongwang@hotmail.com, stsljh@mail.sysu.edu.cn,\\ sdliangyun@163.com,
bianshan@scau.edu.cn, chenyu204@scau.edu.cn
}

\maketitle

\begin{abstract}
Support vector clustering (SVC) is a versatile clustering technique that is able to identify clusters of arbitrary shapes by exploiting the kernel trick. However, one hurdle that restricts the application of SVC lies in its sensitivity to the kernel parameter and the trade-off parameter. Although many extensions of SVC have been developed, to the best of our knowledge, there is still no algorithm that is able to effectively estimate the two crucial parameters in SVC without supervision. In this paper, we propose a novel support vector clustering approach termed ensemble-driven support vector clustering (EDSVC), which for the first time tackles the automatic parameter estimation problem for SVC based on ensemble learning, and is capable of producing robust clustering results in a purely unsupervised manner. Experimental results on multiple real-world datasets demonstrate the effectiveness of our approach.
\end{abstract}

\section{Introduction}

Support vector clustering (SVC) is a flexible clustering technique which is inspired by support vector machines (SVM) \cite{Ben-Hur2001}. In SVC, the input data points are first mapped from the original space to a high-dimensional kernel space, where a sphere that encloses most of the data points is constructed. When mapped back to the original space, the sphere is split into several components, each of which encloses a set of data points and forms a cluster in the clustering result. With the nonlinear mapping, SVC has two major advantages over the other clustering techniques. First, it is capable of identifying clusters of arbitrary shapes. Second, the number of clusters can be obtained automatically in SVC and needn't to be specified in advance.

Ben-Hur et al. \cite{Ben-Hur2001} for the first time introduced the SVC technique, which consists of two phases, i.e., sphere construction and cluster labeling. As both of the two phases are time-consuming, some efforts have been made to improve the efficiency of SVC in recent years. To speedup the sphere construction process, Wang et al. \cite{isvc11} proposed the incremental support vector clustering (ISVC) approach which is able to compute the support vectors (SVs) incrementally and construct the sphere efficiently. Huang et al. \cite{huang2012incremental} extended ISVC by introducing an outlier detection mechanism based on dynamical management of bounded support vectors (BSVs). To improve the efficiency of cluster labeling, Lee and Lee \cite{Lee2005} exploited the topological property of the trained kernel radius function to assign cluster labels. Ping et al. \cite{ping15_KAIS} introduced an adaptive labeling strategy that decomposes clusters into convex hulls, and proposed a fast and scalable SVC (FSSVC) approach. Besides the efficiency, another focus of the SVC research is on its applications. Wang et al. \cite{svstream13} utilized a SVC-based approach to solve the data stream clustering problem. Boecking et al. \cite{Boecking2014} combined SVC with a triangular alignment kernel and used it for time series clustering. Sun et al. \cite{Sun2014} developed a SVC-based model to describe clusters of images with manually tagged words, which is further exploited to deal with automatic or semi-automatic image annotation problems.

Although many SVC based approaches have been developed in recent years \cite{Boecking2014,huang2012incremental,Lee2005,ping15_KAIS,Sun2014,isvc11,svstream13}, most of the existing approaches, if not all, still suffer from a common drawback, i.e., the difficulty in selecting \emph{proper} parameters. Typically, there are two parameters in SVC, namely, the kernel parameter and the trade-off parameter. The kernel parameter decides how data points are mapped to high-dimensional space and thereby controls the shapes of cluster contours in the data space. The trade-off parameter adjusts the generation of bounded support vectors (BSVs) which significantly affects the ability for SVC to deal with noisy data. These two parameters together impose a great influence on the clustering performance of SVC. However, the existing SVC approaches \cite{Boecking2014,huang2012incremental,Lee2005,ping15_KAIS,Sun2014,isvc11,svstream13} generally lack the ability to automatically estimate these two parameters and need to tune the parameters in a supervised manner, i.e., implicitly or explicitly need access to the ground-truth knowledge. To partially address this problem, Wang and Lai \cite{PSVC13} proposed a position regularized support vector clustering (PSVC) approach, which eliminates the selection of the trade-off parameter, yet is still unable to select the kernel parameter effectively and unsupervisedly. It remains a very challenging and unsolved problem how to automatically and properly select the two parameters for SVC in an unsupervised manner.

To tackle this problem, in this paper, we propose an ensemble-driven support vector clustering (EDSVC) approach, which, to the best of our knowledge, is the first algorithm that is capable of automatically estimating the aforementioned two crucial parameters for SVC in a purely unsupervised manner. Our approach is based on the ensemble clustering technique \cite{Fred05_EAC,huang15_NEUCOM,Huang2016,Huang16_TKDE,huang_neucom16,liu2015spectral,Mimaroglu12_diclens,Mimaroglu11_pr,Mimaroglu13_eaai,wu2013theoretic,yi_icdm12}, which has proved to be a powerful tool to accumulate information from multiple (weak) clusterings.
The ground-truth can be viewed as an expert, while the ensemble of multiple clusterings can be viewed as a crowd a individuals. Without the expert, we appeal to the wisdom of the crowd and learn the parameters under the guidance of a crowd of individual clusterings. By exploiting an ensemble learning-based strategy, our approach can automatically estimate the kernel parameter and the trade-off parameter and thereby obtain robust clustering results without supervision. Experimental results on multiple real-world datasets show the effectiveness of the proposed EDSVC approach.

The rest of this paper is organized as follows. The proposed EDSVC algorithm is introduced in Section~\ref{sec:EDSVC}. The experimental results are reported in Section~\ref{sec:experiment}. Finally, we conclude this paper and discuss the future work in Section~\ref{sec:conclusion}.

\section{Proposed Framework}
\label{sec:EDSVC}

In this section, we introduce the proposed clustering framework. First, we introduce the SVC process in Section~\ref{sec:SVC}. Then, in Section~\ref{sec:ensemble_eval}, we describe the generation of an ensemble of multiple clusterings, which is further exploited to guide the parameter estimation process. Finally, we summarize the overall algorithm of EDSVC in Section~\ref{sec:overall_algo}.

\subsection{Support Vector Clustering}
\label{sec:SVC}

Let $\mathcal{X}=\{x_1,\cdots,x_N\}$ be a data set, where $x_i\in\mathbb{R}^d$ is the $i$-th data point and $N$ is the number of data points in $\mathcal{X}$. By a nonlinear transformation $\Phi$, the data points in $\mathcal{X}$ are mapped to a high-dimensional space and the smallest enclosing sphere with radius $R$ is found. That is
\begin{equation}
\label{eq:smallest_sphere} \|\Phi(x_i)-\mu\|^2=R+\xi_i, \forall
i=1,\dots,N,
\end{equation}
where $\mu$ is the center of the sphere and $\xi_i \geq 0$, for
$i=1,\dots,N$, are the slack variables which incorporates soft
constraints.

To solve the problem in Eq.(\ref{eq:smallest_sphere}), the Lagrangian is introduced as follows:
\begin{eqnarray}
\label{eq:lagrangian}
&L=&R^2-\sum_{i}{(R^2+\xi_i-\|\Phi(x_i)-\mu\|^2)\beta_i}\nonumber\\
&&-\sum{\xi_i \alpha_i}+C\sum{\xi_i},
\end{eqnarray}
where $\beta_i\geq 0$ and $\alpha_i \geq 0$ are Lagrange
multipliers, $C$ is a constant and $C\sum{\xi_i}$ is the penalty
term. By setting the derivative of $L$ w.r.t. $R$, $\mu$, and
$\xi_i$, respectively, to zero, we have
\begin{equation}
\label{eq:derivative}
\sum_i{\beta_i}=1,~~~\mu=\sum_i{\beta_i\Phi(x_i)},~~~\beta_i=C-\alpha_i.
\end{equation}
The KKT complementary conditions result in
\begin{equation}
\label{KKT} \xi_i \alpha_i=0,~~~
(R^2+\xi_i-\|\Phi(x_i)-\mu\|^2)\beta_i=0.
\end{equation}
Then, we eliminate the variables $R$, $\mu$, $\xi_i$ and $\alpha_i$, and
turn the Lagrangian into the Wolfe dual form. That is
\begin{eqnarray}
\label{eq:wolfe}
\max_{\beta_i}W=\sum_i{K(x_i,x_i)\beta_i}-\sum_{i,j}{\beta_i\beta_jK(x_i,x_j)}\\
s.t. ~~~\sum_i{\beta_i}=1,~0\leq\beta_i\leq C,\forall i=1,\dots,N,\nonumber
\end{eqnarray}
where the dot product $\Phi(x_i)\cdot \Phi(x_j)$ is represented by
the Gaussian kernel $K(x_i,x_j)=\exp(-q\|x_i-x_j\|^2)$ with width
parameter $q$.

With regard to the values of the Lagrangian multipliers $\beta_i$, there are two types of representative points. The points with $0<\beta_i<C$ are termed support vectors (SVs) which lie on the sphere surface. The points with $\beta_i=C$ are termed bounded support vectors (BSVs) which lie outside the sphere and are usually treated as outliers. Obviously, there will be no BSV when setting $C\geq 1$.

In SVC, the cluster boundaries are delineated by exploiting the SVs, and the data points can be assigned with cluster labels by means of various cluster labeling techniques \cite{Ben-Hur2001,Lee2005}. After cluster labeling, the final clustering result of SVC can be obtained. Formally, we denote the function of the SVC process as follows:
\begin{equation}
\pi^*_{(q,C)} = SVC(\mathcal{X},q,C),
\end{equation}
where $\mathcal{X}$ is the input data, $q$ is the kernel parameter, $C$ is the trade-off parameter, and $\pi^*_{(q,C)}$ is the clustering result by SVC with parameters $q$ and $C$.

\subsection{Ensemble Generation}
\label{sec:ensemble_eval}

In the existing SVC-based approaches \cite{Boecking2014,huang2012incremental,Lee2005,ping15_KAIS,Sun2014,isvc11,svstream13}, the kernel parameter and the trade-off parameter are generally selected by some \emph{trial and error} strategy, which explicitly or implicitly needs access to the ground-truth knowledge. Different from these methods, in this paper, we propose to automatically estimate the parameters for SVC in a purely unsupervised manner. Instead of using ground-truth knowledge, we take advantage of the ensemble clustering technique and estimate parameters with the help of a diverse set of clusterings.

Ensemble clustering is a clustering technique that aims to combine multiple clusterings into a probably better and more robust clustering \cite{Fred05_EAC,huang15_NEUCOM,Huang2016,Huang16_TKDE,liu2015spectral,Mimaroglu12_diclens,Mimaroglu11_pr,Mimaroglu13_eaai,wu2013theoretic,yi_icdm12}. Each of the input clusterings is referred to as a base clustering, which in fact can be generated by any clustering algorithms. In this paper, we generate the ensemble of multiple base clusterings by $k$-means. For the generation of each base clustering, the number of clusters for $k$-means is randomly selected in the interval of $[2, \sqrt[3]{N}]$, where $N$ is the number of data points in the dataset $\mathcal{X}$. Formally, the ensemble of base clusterings is denoted as
\begin{equation}
\Pi=\{\pi^1,\cdots,\pi^M\},
\end{equation}
where $\pi^i$ denotes the $i$-the base clustering in the ensemble $\Pi$, and $M$ denotes the number of base clusterings in $\Pi$. Without loss of generality, in this paper, we use $M=10$ base clusterings. Each base clustering is a partition of the original dataset. With the ensemble generated by random initializations and random number of clusters, a single base clustering (i.e., an ensemble member) may be unstable (or low-quality), but the crowd of diverse members can provide different views of the data and may together provide reliable information about the inherent cluster structures. We argue that the ensemble of multiple clusterings can form a robust guidance for the process of unsupervised parameter estimation, which will be described in detail and demonstrated by extensive experiments in the following sections of this paper.

\subsection{Overall Algorithm}
\label{sec:overall_algo}

In this paper, we aim to estimate the kernel parameter and the trade-off parameter with the help of an ensemble of multiple clusterings and thereby to achieve robust clustering results in an unsupervised manner. Our work is partially inspired by the idea of ``the wisdom of crowds'', which is the process of taking into consideration the opinion of a crowd of individuals rather than a single expert in the field of economic and social science. In our approach, without ground-truth labels, we appeal to the crowd of multiple base clusterings to assess the clustering performance of SVC with different parameter settings. Specifically, we evaluate the quality of the output clusterings generated by SVC using different parameters $q$ and $C$, and aim to find the best parameters automatically. The first task here is to define an unsupervised criterion to evaluate the output clusterings.

The normalized mutual information (NMI) \cite{strehl02} provides a sound indication of the shared information between two clusterings and is widely used for evaluating the quality of clusterings. Generally, the NMI score is computed between a test clustering and the ground-truth clustering so as to evaluate the quality of the test clustering. However, in the unsupervised setting of this work, the ground-truth information is not available. Instead of using the ground-truth, we propose to evaluate the quality of a candidate clustering with the help of an ensemble of multiple clusterings. Let $\pi^*_{(q,C)} = SVC(\mathcal{X},q,C)$ be a candidate clustering generated by SVC with the kernel parameter $q$ and the trade-off parameter $C$. The average NMI score is computed between the clustering $\pi^*_{(q,C)}$ and the ensemble $\Pi$. That is
\begin{equation}
\label{eq:ANMI}
ANMI(\pi^*_{(q,C)},\Pi)=\frac{1}{M}\sum_{\pi^i\in\Pi}NMI(\pi^i,\pi^*_{(q,C)}),
\end{equation}
where $NMI(\pi^i,\pi^*_{(q,C)})$ denotes the NMI score between two clusterings $\pi^i$ and $\pi^*_{(q,C)}$, and $M$ is the number of base clusterings in $\Pi$.

With the evaluation criterion defined in Eq.~(\ref{eq:ANMI}), we can evaluate the clustering results produced by SVC with different parameters. Specifically, we aim to find the optimal parameters $\hat{q}$ and $\hat{C}$ that maximize the average NMI between the ensemble $\Pi$ and the clustering result produced by SVC. That is
\begin{equation}
\label{eq:object_qC}
(\hat{q},\hat{C})={\arg\max}_{(q,C)} ANMI(\pi^*_{(q,C)},\Pi).
\end{equation}

In this paper, we solve the optimization problem in Eq.~(\ref{eq:object_qC}) by evaluating a set of candidate values. Let $\mathcal{Q}=\{q_1,\cdots,q_{n_q}\}$ be a set of $n_q$ candidate values for the parameter $q$. Let $\mathcal{C}=\{C_1,\cdots,C_{n_c}\}$ be a set of $n_c$ candidate values for the parameter $C$. In this paper, we use $n_q=100$ values for $q$ and $n_c=100$ values for $C$. Of the two parameters $q$ and $C$, one parameter is optimized by fixing the other one, and vice versa. First, by fixing the parameter $C=C_0$, we test the candidate values in $\mathcal{Q}$ for the parameter $q$ and find the best value $\hat{q}$ that optimized Eq.~(\ref{eq:object_qC}). Then, by fixing the parameter $q=\hat{q}$, we test the candidate values in $\mathcal{C}$ for the parameter $C$ and find the optimal value $\hat{C}$. With the optimal parameters obtained by Eq.~(\ref{eq:object_qC}), the final clustering result is generated by SVC using the estimated parameters $\hat{q}$ and $\hat{C}$.

For clarity, the proposed EDSVC algorithm is summarized in Algorithm 1.

\begin{figure}[!h]
\textbf{Algorithm 1 (Ensemble-Driven Support Vector Clustering)}\\
\small{ {\bfseries Input:} $\mathcal{X}=\{x_1,\cdots,x_N\}$
\begin{algorithmic}[1]
    \STATE Generate the ensemble $\Pi=\{\pi^1,\cdots,\pi^M\}$:\\
    \textbf{for} {$i=1,2,\cdots,M$}\\
     ~~~~Select $k$ randomly in the interval of $[2, \sqrt[3]{N}]$.\\
     ~~~~Build the $i$-th base clustering $\pi^i$ using $k$-means.\\
    \textbf{end for}
    \STATE Unsupervised parameter estimation:\\
    ~~~~By fixing $C=C_0$, find the optimal $\hat{q}$ in $\mathcal{Q}$ w.r.t. Eq.~(\ref{eq:object_qC}).\\
    ~~~~By fixing $q=\hat{q}$, find the optimal $\hat{C}$ in $\mathcal{C}$ w.r.t. Eq.~(\ref{eq:object_qC}). \\
    \STATE Obtain the clustering result by SVC using the estimated parameters $\hat{q}$ and $\hat{C}$:\\
    ~~~~$\tilde{\pi} = SVC(\mathcal{X},\hat{q},\hat{C})$
\end{algorithmic}
{\bfseries Output:} the final clustering $\tilde{\pi}$.}
\end{figure}

\section{Experiments}
\label{sec:experiment}

In this section, we conduct experiments on multiple real-world datasets to evaluate the performance of the proposed EDSVC approach. We first introduce the datasets and the evaluation metric in Section~\ref{sec:dataset_eval}. Then we compare EDSVC against the base clusterings in the ensemble $\Pi$, against four ensemble clustering methods, and against position regularized support vector clustering (PSVC) \cite{PSVC13}, in Section~\ref{sec:comp_base}, Section~\ref{sec:comp_ensemble}, and Section~\ref{sec:comp_psvc}, respectively.

\subsection{Datasets and Evaluation Metric}
\label{sec:dataset_eval}

\begin{table}[!t]\vskip -0.1in
\centering
\caption{Description of datasets}
\label{table:datasets}
\begin{center}
\begin{tabular}{p{2cm}<{\centering}|p{1.2cm}<{\centering}p{1.2cm}<{\centering}p{1.2cm}<{\centering}}
\toprule
Dataset         &\#Object &\#Attribute   &\#Class \\
\midrule
\emph{Wine}    &178  &13    &3\\
\emph{Ionosphere}    &351    &34     &2\\
\emph{BC}       &683    &9     &2\\
\emph{BA}     &1,372    &4      &2\\
\emph{Yeast}   &1,484  &8    &10\\
\emph{Semeion}  &1,593  &256    &10\\
\bottomrule
\end{tabular}
\end{center}\vskip -0.3in
\end{table}

In this paper, the experiments are conducted on six real-world datasets, namely, \emph{Wine}, \emph{Ionosphere}, \emph{Breast Cancer} (\emph{BC}), \emph{Banknote Authentication} (\emph{BA}), \emph{Yeast}, and \emph{Semeion}. The six datasets are publicly available at the UCI machine learning repository \cite{Bache+Lichman:2013}. The details of the benchmark datasets are given in Table~\ref{table:datasets}.

In the experiments, the normalized mutual information (NMI) \cite{strehl02} is used to evaluate the quality of clusterings. The NMI score provides a sound indication of the shared information between clusterings. A higher NMI indicates a better clustering result.

\subsection{Comparison Against Base Clusterings}
\label{sec:comp_base}

\begin{figure}[!b]\vskip -0.1in
\begin{center}
{
{\includegraphics[width=0.9\columnwidth]{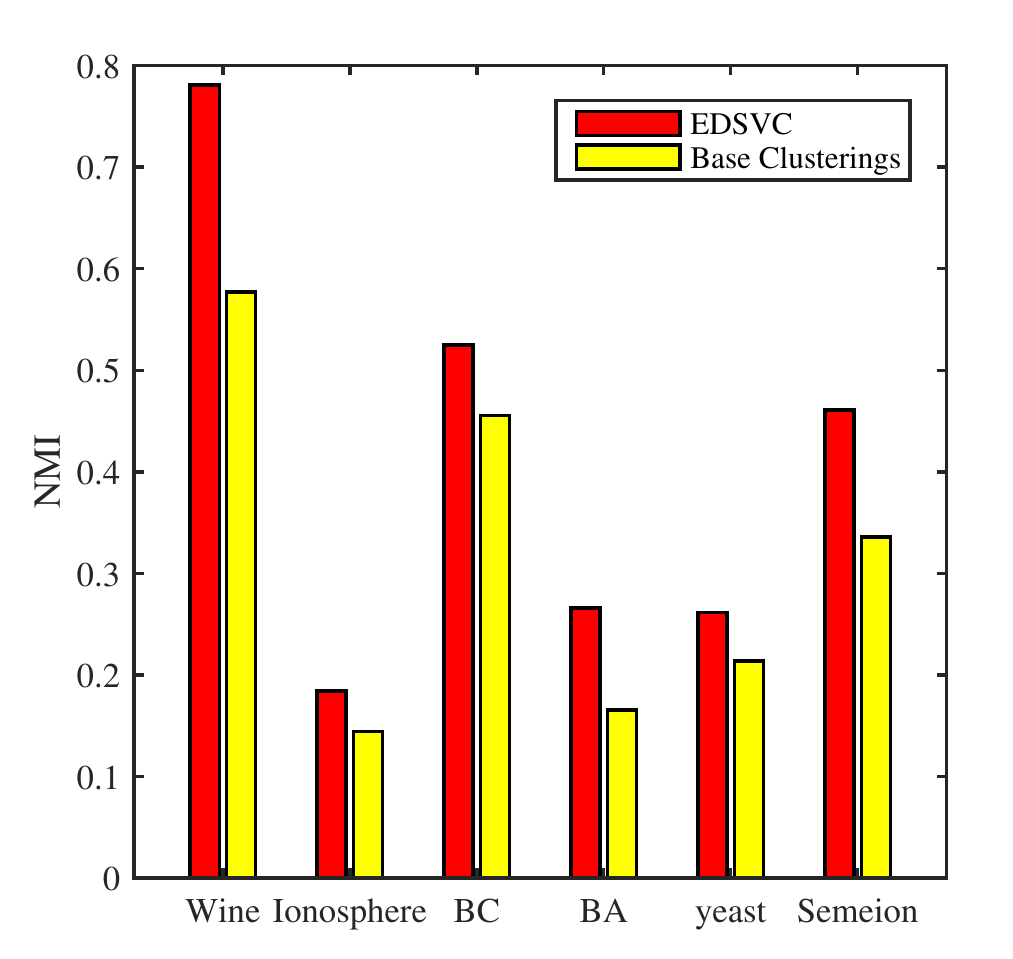}}}
\caption{Comparing EDSVC with the base clusterings in terms of NMI.}\vskip -0.1in
\label{fig:comp_base}
\end{center}\vskip -0.3in
\end{figure}

In this paper, we propose the EDSVC approach which bridges the gap between ensemble clustering and support vector clustering. For each benchmark dataset, an ensemble of multiple base clusterings is exploited to guide the SVC process. The purpose is to utilize the information of multiple clusterings to obtain a probably better and more robust clustering result. In this section, we compare the clustering results of EDSVC against the average performance of the base clusterings. As can be seen in Fig.~\ref{fig:comp_base}, the EDSVC approach is able to produce significantly better clustering results (in terms of NMI) than the base clusterings on all of the six benchmark datasets.

\subsection{Comparison Against Ensemble Clustering Methods}
\label{sec:comp_ensemble}

\begin{table*}[!t]\vskip -0.25in
\centering
\caption{The performance (in terms of NMI) of EDSVC compared against the ensemble clustering approaches. The best score in each column is highlighted in bold.}\vskip -0.19in
\label{table:comp_auto}
\begin{center}
\begin{tabular}{p{4cm}<{\centering}|p{1.4cm}<{\centering}p{1.4cm}<{\centering}p{1.4cm}<{\centering}p{1.4cm}<{\centering}p{1.4cm}<{\centering}p{1.4cm}<{\centering}}
\toprule
Method         &Wine &Ionosphere   &BC          &BA  &Yeast    &Semeion\\
\midrule
EDSVC        &\textbf{0.781}       &\textbf{0.184}       &\textbf{0.525}       &\textbf{0.266}       &\textbf{0.267}       &\textbf{0.461}\\
\midrule
COMUSA \cite{Mimaroglu11_pr}        &0.580       &0.080       &0.219       &0.128       &0.150       &0.122\\
DICLENS \cite{Mimaroglu12_diclens}        &0.611       &0.067       &0.369       &0.205       &0.184       &0.326\\
COMUSACL \cite{Mimaroglu13_eaai}        &0.671       &0.164       &0.480       &0.232       &0.204       &0.373\\
COMUSACL-DEW \cite{Mimaroglu13_eaai}        &0.703       &0.134       &0.494       &0.212       &0.264       &0.344\\
\bottomrule
\end{tabular}
\end{center}\vskip -0.2in
\end{table*}

\begin{figure*}[!t]
\begin{center}
{\subfigure[\emph{Wine}]
{\includegraphics[width=0.59\columnwidth]{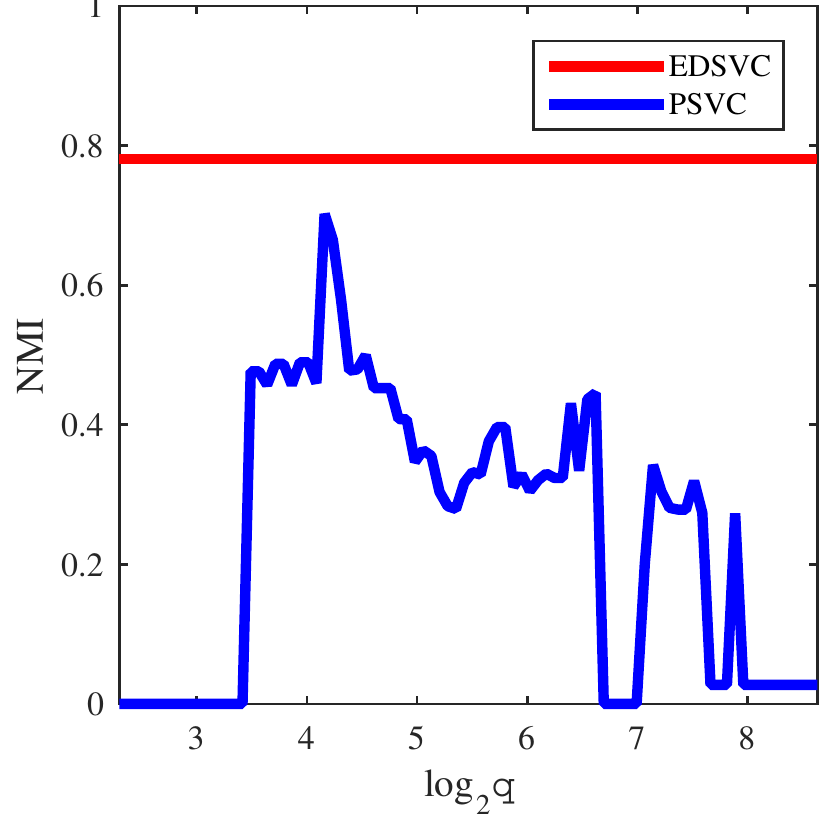}}}
{\subfigure[\emph{Ionosphere}]
{\includegraphics[width=0.59\columnwidth]{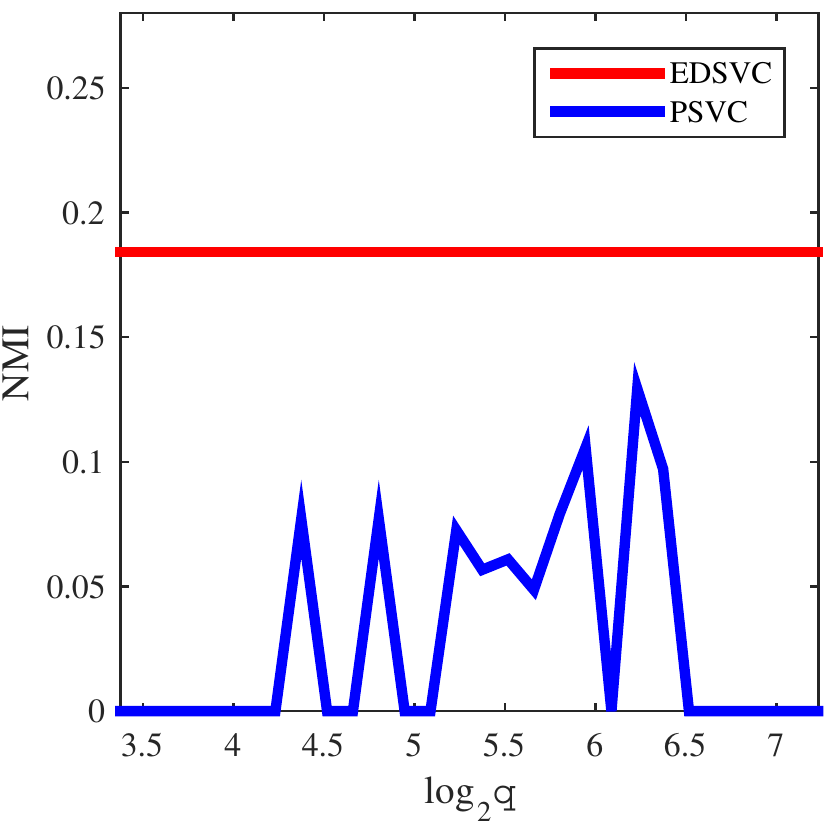}}}
{\subfigure[\emph{BC}]
{\includegraphics[width=0.59\columnwidth]{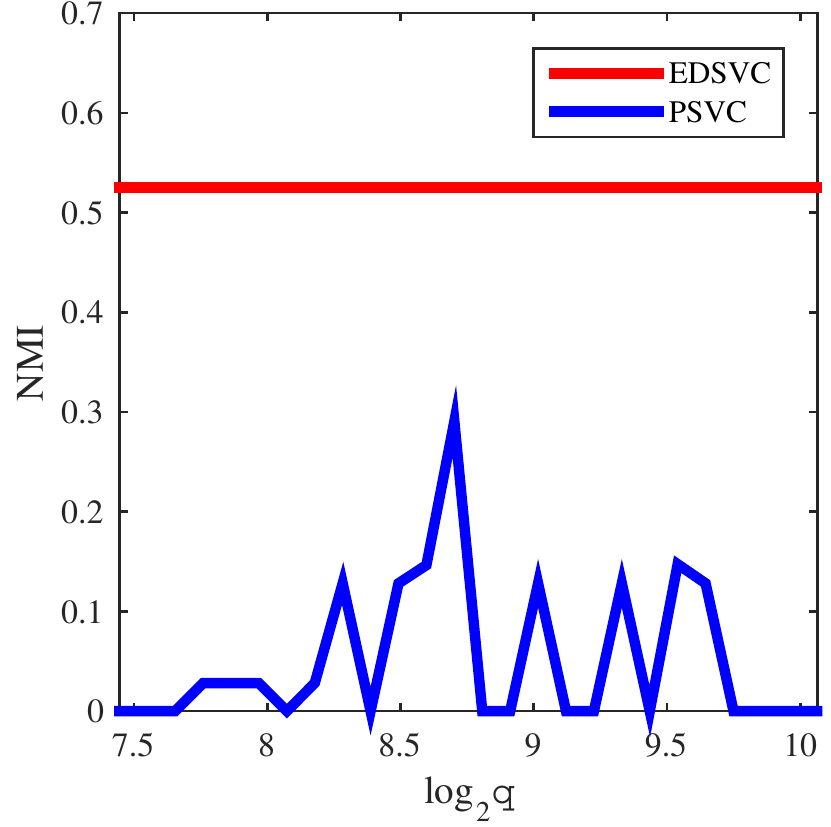}}}
{\subfigure[\emph{BA}]
{\includegraphics[width=0.59\columnwidth]{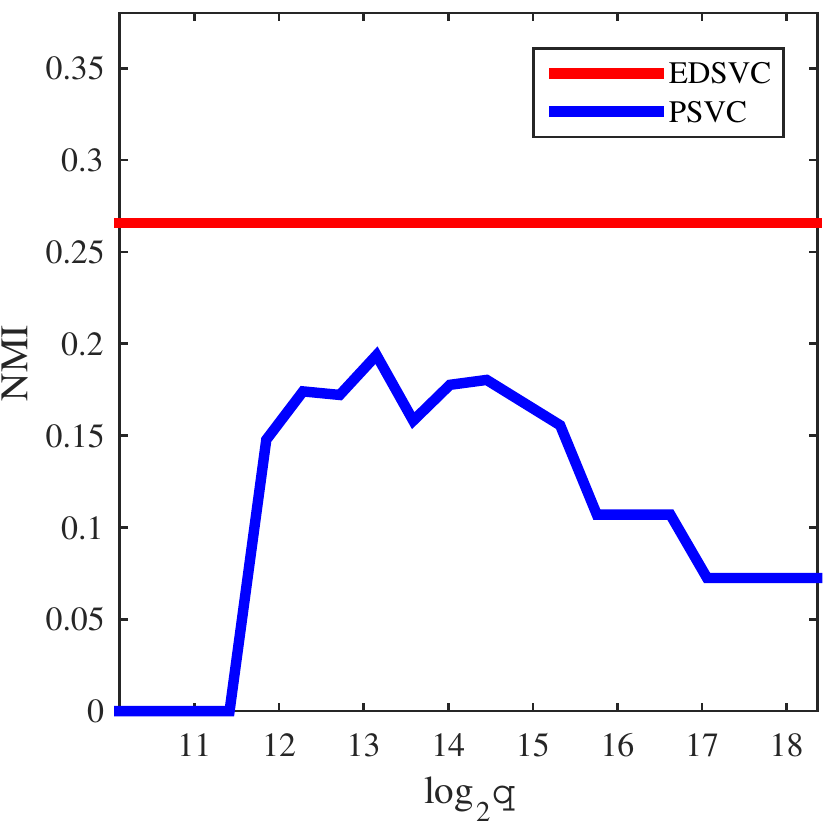}}}
{\subfigure[\emph{Yeast}]
{\includegraphics[width=0.59\columnwidth]{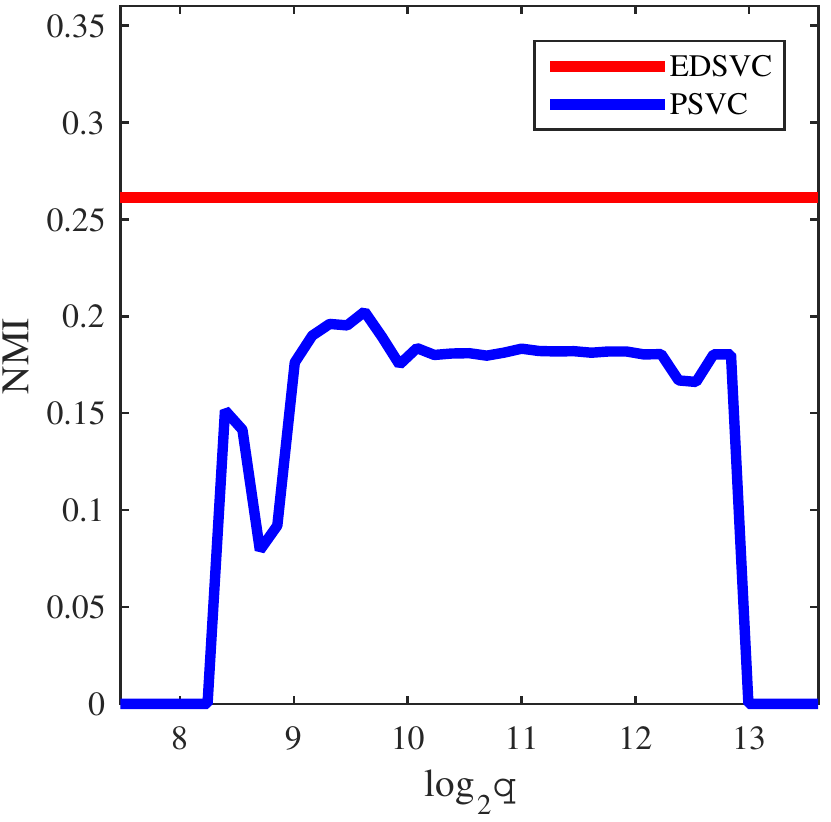}}}
{\subfigure[\emph{Semeion}]
{\includegraphics[width=0.59\columnwidth]{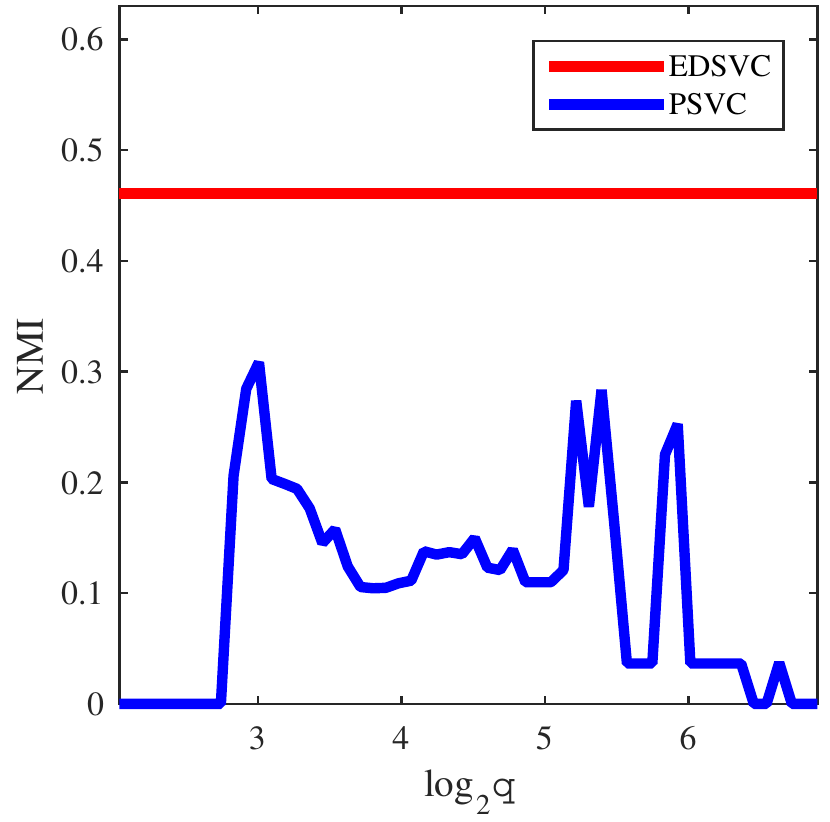}}}\vskip -0.15in
\caption{Comparing the performances of EDSVC and PSVC \cite{PSVC13} (with varying parameter $q$) on the benchmark datasets. Note that the X axis corresponds to $\log_2q$.}
\label{fig:en_sizes}
\end{center}\vskip -0.4in
\end{figure*}

In this section, we compare EDSVC against four state-of-the-art ensemble clustering approaches, namely, COMUSA \cite{Mimaroglu11_pr}, DICLENS \cite{Mimaroglu12_diclens}, COMUSACL \cite{Mimaroglu13_eaai}, and COMUSACL-DEW \cite{Mimaroglu13_eaai}. To provide a fair comparison, on each benchmark dataset, the same ensemble of clusterings are used for both EDSVC and the ensemble clustering approaches. The performances of EDSVC and the four baseline approaches in terms of NMI are reported in Table~\ref{table:comp_auto}. Our EDSVC approach outperforms the baseline ensemble clustering approaches on all of the six benchmark datasets. Especially, for the \emph{Wine}, \emph{Ionosphere}, \emph{BA}, and \emph{Semeion} datasets, our approach yields significantly better performance than the baseline approaches.

\subsection{Comparison Against PSVC}
\label{sec:comp_psvc}

In this section, we further compare EDSVC against the position regularized support vector clustering (PSVC) approach \cite{PSVC13}. The PSVC approach eliminates the trade-off parameter $C$, but is still unable to automatically estimate the kernel parameter $q$. In comparison, our EDSVC approach is capable of automatically estimating $q$ and $C$ and producing robust clustering results. Because PSVC cannot determine the parameter $q$ automatically, we illustrate the NMI scores of our EDSVC and that of PSVC w.r.t. varying $q$ in Fig.~\ref{fig:en_sizes}. As can be seen in Fig.~\ref{fig:en_sizes}, the clustering performance of PSVC is sensitive to the kernel parameter $q$. Even when the best $q$ is selected for PSVC on each benchmark dataset, our EDSVC approach can still produce consistently better clustering results than PSVC, which demonstrates the effectiveness of the proposed unsupervised parameter selection strategy and the robustness of the proposed EDSVC approach.

\section{Conclusion and Future Work}
\label{sec:conclusion}

In this paper, we propose a novel support vector clustering (SVC) approach termed ensemble-driven support vector clustering (EDSVC), which for the first time tackles the problem of unsupervised parameter estimation for SVC based on the ensemble clustering technique. Without needing access to the ground-truth, our approach is able to automatically estimate the kernel parameter $q$ and the trade-off parameter $C$ under the guidance of an ensemble of multiple base clusterings. We conduct experiments on six real-world datasets. The experimental results have demonstrated the superiority of our approach over other ensemble-based and SVC-based approaches. In the future work, we plan to develop a scalable version of EDSVC and enable it to handle the clustering problem of large-scale datasets.

\section{Acknowledgement}
This work was supported by NSFC (61573387 \& 61502543),
Guangdong Natural Science Funds for Distinguished Young Scholar
(2016A030306014), the PhD Start-up Fund of Natural Science Foundation
of Guangdong Province, China (2016A030310457, 2015A030310450 \&
2014A030310180), the Science and Technology Planning Project of Guangdong Province (2016A050502050 \& 2015A020209124), the Fundamental Research Funds for the Central
Universities (16lgzd15), and SCAU
Special Funds for Young Scientific Talents.

\nocite{ex1,ex2}
\bibliographystyle{latex12}
\bibliography{icpr2016}

\end{document}